\pdfoutput=1
\documentclass[runningheads]{llncs}
\usepackage[T1]{fontenc}
\usepackage{graphicx}
\usepackage{amsmath}
\usepackage{csquotes}

\renewcommand{\thefigure}{\arabic{figure}}

\begin{document}
\title{SwarmChat: An LLM-Based, Context-Aware Multimodal Interaction System for Robotic Swarms\thanks{This paper has been accepted and presented at the 16th International Conference on Swarm Intelligence (ICSI 2025), held on July 11--15, 2025, in Yokohama, Japan.}} 
\titlerunning{SwarmChat}
\author{Ettilla Mohiuddin Eumi\inst{1}\orcidID{0009-0000-7001-9338} \and Hussein Abbass\inst{1}\orcidID{0000-0002-8837-0748} \and Nadine Marcus\inst{2}\orcidID{0000-0002-4996-4850}}
\authorrunning{EM. Eumi et al.}
\institute{School of Systems \& Computing, UNSW Canberra, Canberra ACT 2600, Australia\\
\email{\{e.eumi, h.abbass\}@unsw.edu.au} \and School of Computer Science and Engineering, UNSW Sydney, Sydney NSW 2052, Australia\\ \email{nadinem@unsw.edu.au}}

\maketitle

\begin{abstract}
Traditional Human-Swarm Interaction (HSI) methods often lack intuitive real-time adaptive interfaces, making decision making slower and increasing cognitive load while limiting command flexibility. To solve this, we present SwarmChat, a context-aware, multimodal interaction system powered by Large Language Models (LLMs). SwarmChat enables users to issue natural language commands to robotic swarms using multiple modalities, such as text, voice, or teleoperation. The system integrates four LLM-based modules: Context Generator, Intent Recognition, Task Planner, and Modality Selector. These modules collaboratively generate context from keywords, detect user intent, adapt commands based on real-time robot state, and suggest optimal communication modalities. Its three-layer architecture offers a dynamic interface with both fixed and customisable command options, supporting flexible control while optimising cognitive effort. The preliminary evaluation also shows that the SwarmChat\textquoteright s LLM modules provide accurate context interpretation, relevant intent recognition, and effective command delivery, achieving high user satisfaction. 
\end{abstract}

\section{Introduction}
The term \enquote{Multimodal Interaction} refers to a transformative approach to Human-Computer Interaction (HCI) that enables users to engage with systems through various communication channels, such as speech, gesture, touch, and gaze~\cite{dritsas2025multimodal}. 
In HCI, the concepts of \enquote{Human Swarm} and \enquote{Human Swarm Teaming} have gained increasing interest among researchers, who are focused on how humans can effectively control and collaborate with groups of robots or agents. Swarm intelligence describes the collective behaviour of simple individuals working together to solve complex problems. This phenomenon is commonly observed in nature, particularly among ants and bees. These systems rely on self-organisation and decentralised coordination, where individuals interact locally and communicate without needing a central leader, ultimately leading to efficient problem solving through local interactions~\cite{kordon2010swarm}.

The concept of \enquote{Human Swarm} is a recent development designed to enable large groups of participants to collaborate in real-time to (a) process noisy information, (b) weigh different options, and (c) converge on final decisions through a competitive process among various subgroups~\cite{rosenberg2015human}. Additionally, Human-Swarm Teaming (HST) is a growing field that focuses on operating and controlling large groups of autonomous or semi-autonomous robot swarms with assistance from human operators. Some scholars advocate for scalable swarm robotics in this research area, in which numerous low-cost robots exhibit collective behaviour. This approach enables human operators to manage large numbers of robots with minimal input by adjusting key parameters that influence overall swarm behaviour~\cite{pendleton2013human}.

However, a significant challenge is that robotic swarms are not fully autonomous, making effective human-swarm interaction crucial for deploying these systems in real-world environments. Human operators possess a higher understanding of the algorithms driving swarm intelligence, and their involvement is essential for solving real-world problems in conjunction with robot swarms~\cite{fam2023human}. Moreover, it is essential to ensure that human operators can intuitively and effectively communicate with these robotic swarms.
In the context of human-swarm interaction, it is necessary to design an interaction framework that (a) ensures clear and understandable communication between humans and swarms, (b) provides a dynamic interface that displays real-time data, and (c) supports quick and efficient decision-making through an intuitive interface.
Furthermore, multimodality enables systems to mimic the natural way users prefer to communicate with other humans and with their environment. Incorporating multiple sensory channels into human-machine communication ensures adaptive interaction and greater flexibility in HCI~\cite{dritsas2025multimodal}.

Our developed \emph{SwarmChat} addresses these challenges by leveraging the robust capabilities of large language models (LLMs) and context-aware computing to manage the complex dynamics of robotic swarms. This system utilises multimodal interactions, such as text, voice, and teleoperation, and integrates real-time analytics to adapt to user input and environmental changes effectively. Moreover, by incorporating a sophisticated natural language processing engine, \emph{SwarmChat} can interpret complex candidate commands and convert them into actionable tasks easily understood by the swarms.
The primary objectives of \emph{SwarmChat} include:
\begin{itemize}
    \item Development of an intuitive communication protocol that allows users to issue commands in natural language through multi-modalities. 
    \item Implement a real-time adaptive interface that displays relevant swarm data and status updates, giving the users real-time dynamic updates.
    \item Establishment of a quick decision-making environment to provide a stress-free command selection.
\end{itemize}

The structure of this paper is as follows: \textbf{Section 2} reviews the relevant literature on Human Swarm Interaction. \textbf{Section 3} outlines the system architecture \& overall framework of \emph{SwarmChat}. \textbf{Section 4} addresses context awareness and cognitive load management. \textbf{Section 5} system evaluation and LLM module analysis and \textbf{Section 6} conclusion \& future work.

\section{Related Research}
Significant progress has been made in Human-Swarm Interaction, with researchers exploring various communication media to facilitate seamless human-swarm collaboration. For example, one study showed that autonomous robots can be coordinated in real time using predefined \enquote{plays} and dynamic \enquote{audibles} in a simulated search-and-rescue mission, where verbal and nonverbal cues proved effective~\cite{chaffey2020human}.

Another multichannel HSI system integrated 3D gestures and natural language to enhance command understanding and user experience, though limited by gesture sensor range and robot design requirements~\cite{chen2020multichannel}. Drone swarms have been used to communicate emotion and intention through motion-based nonverbal patterns, enhancing communicative capacity, though lacking audio integration~\cite{grispino2020evaluating}. The \enquote{MICAH} framework highlights five key aspects for adaptive interaction: mission goals, complexity, autonomy levels, interaction dynamics, and human states, emphasizing the need for multimodal data fusion to assess cognitive load~\cite{hussein2022characterization}. \emph{FlockGPT} introduced natural language control of drone formations using generative AI and Signed Distance Function geometry, addressing interaction interface gaps~\cite{lykov2024flockgpt}.

Sequential hand gesture recognition has enabled decentralised swarm control in virtual and physical environments, reducing cognitive load but requiring better accuracy and reinforcement learning~\cite{kakish2021towards}. \emph{JSwarm}, a Jingulu-inspired language, enables bidirectional communication with swarms to reduce ambiguity but needs further real-world usability testing~\cite{abbass2022jswarm}. Gesture-based communication in underwater HRI has been explored to improve diver-AUV collaboration amid environmental challenges~\cite{aldhaheri2024underwater}. The DVRP-MHSI platform allows intuitive, multimodal control of mobile robots using low-cost devices, supporting real-time visualisation and reducing operator workload~\cite{zhu2024dvrp}.

In aquatic HSI, pupillometry and brain activity were used to measure cognitive load and model object recognition~\cite{bhattacharya2022designing}. Another work employed the “MICAH” model to optimize teaming by mapping automation and human state dynamics. Personalised human-swarm interaction was another area of focus, where a machine-learning-based Body-Machine Interface (BoMI) adapted to individual users, with hand movements via \enquote{LEAP} Motion emerging as the most effective control method~\cite{macchini2021personalized}.

After reviewing relevant articles on communication channels, seamless control, and cognitive load, it is evident that multimodal communication in human-swarm interaction is still in its early stages. While some studies have explored gestures and other input modalities, significant gaps remain, particularly in handling the unpredictability of swarm behaviour. Although control algorithms have advanced, human integration is still limited—improving this could enhance swarm predictability and control. Existing platforms often lack the flexibility to support modalities like brain-computer and eye-tracking interfaces. The DVRP-MHSI platform attempts to fill this gap by enabling the flexible testing of swarm algorithms and interaction methods~\cite{zhu2024dvrp}. Studies also reveal that users prefer switching modalities to express intent better and reduce stress~\cite{oviatt2004we} as task complexity increases.
Further research is needed to develop more natural, intuitive, and convenient interfaces for effective human-swarm interaction. 

\section{System Architecture and Overall Framework}
\begin{figure}
\centering
\includegraphics[width=\textwidth]{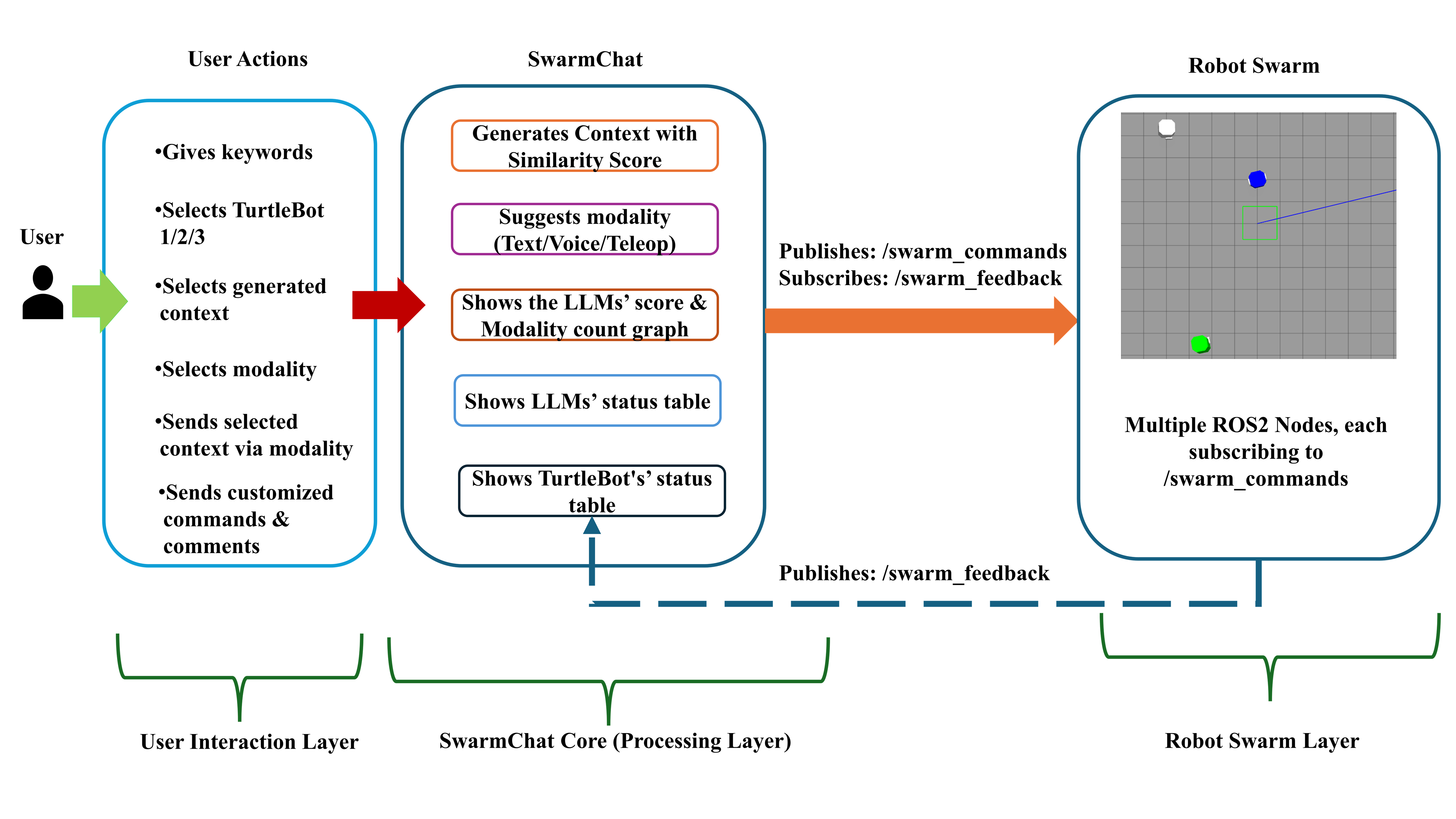}
\caption{The architecture of SwarmChat.}
\label{SwC}
\end{figure}

\emph{SwarmChat} is developed as a comprehensive end-to-end framework that converts natural language commands into detailed instructions for controlling swarms. It integrates context-aware command processing with large language model (LLM) interactions, enabling users to control robot swarms via text, voice, and teleoperation. This multimodal approach is designed to help reduce operators' stress and create a dynamic interface that facilitates seamless communication between human operators and robot swarms. The system is built on a three-layer architecture: the User Interaction Layer, the SwarmChat Core (Processing Layer), and the Robot Swarm Layer.
This layered architecture enables SwarmChat to provide both high-level command abstraction and low-level command execution, while also supporting real-time feedback and continuous user interaction.
The whole architecture is shown in Fig.~\ref{SwC}.

\subsection{User Interaction Layer: }
Users interact with the system through a Tkinter-based graphical user interface (GUI), initially providing raw input keywords (e.g., \textit{K} = \{\enquote{move}, \enquote{forward}, \enquote{patrol}\}). Once submitted, the system generates four candidate command contexts (e.g., \enquote{Go and patrol the area}, \enquote{Move forward and patrol the area}). Each context is evaluated using Jaccard similarity~\cite{niwattanakul2013using}, scaled between 0.6 and 1.0 to ensure even the less relevant contexts retain baseline relevance, supporting balanced LLM decision-making. These options are displayed on the GUI for user review and selection.

After selecting a context, the system suggests a communication modality (Text, Voice, or Teleop), which users can accept or override. Users may also send custom commands with any modality and optionally provide feedback on their customised actions. Finally, the user selects a target robot (e.g., TurtleBot 1, 2, or 3), and the system sends the command for execution.

\subsection{SwarmChat Core (Processing Layer)}
The SwarmChat Core serves as the system\textquoteright s \enquote{brain}, transforming user-provided keywords into refined, context-aware instructions executable by the robot swarm. It integrates multiple LLM-based modules, each responsible for a specific stage in a processing pipeline—from keyword extraction to command packaging—supported by context-awareness mechanisms that adapt commands based on real-time robot states. The final output is a command message and selected communication modality published to the swarm via ROS2. This layer performs the following actions:
\subsubsection{Context Generation:}
The \textbf{Context Generator} module converts the keyword set \textit{K} into four candidate command contexts using the mapping function:
\begin{equation}
f_{CG} : K \rightarrow \{ c_1, c_2, c_3, c_4 \}
\end{equation}

For example, with \textit{K} = \{``move'', ``forward'', ``patrol''\}, generated contexts might include: \(c_1\): Patrol the area, \(c_2\): Move forward and patrol, \(c_3\): Move left and patrol, \(c_4\): Move right and patrol.

\subsubsection{Similarity Scoring Module:}
This module evaluates each candidate context using Jaccard similarity with the user\textquoteright s keywords. Scores are scaled to [0.6, 1.0] to avoid eliminating less similar contexts. For instance, \enquote{patrol area} and \enquote{patrol zone} yield scores of 1.0 and 0.9, respectively. These are shown in the GUI under ``Context Similarity Scores.''

\subsubsection{Intent Recognition:}
This module uses rule-based checks to infer user intent from keyword sets. For example:
\begin{itemize}
    \item ``patrol'' $\rightarrow$ ``Patrol mode activated.''
    \item ``go'' $\rightarrow$ ``Navigation mode activated.''
    \item Otherwise $\rightarrow$ ``General operation.''
\end{itemize}

\subsubsection{Task Planner:}
Incorporates real-time robot state (e.g., location, battery) into the selected context, enhancing context awareness.

\subsubsection{Modality Selection:}
Chooses between Text, Voice, and Teleop based on context content and similarity score: $\ge 0.85$ $\rightarrow$ Teleop; contains ``speak'' $\rightarrow$ Voice; otherwise $\rightarrow$ Text. Users may override suggestions.

\subsubsection{Command Packaging and ROS2 Publication:}
Final commands are packaged and published via ROS2. Confirmation appears in the sender terminal (e.g., ``Published command: ...''). The last command sent by voice may reflect user customization.

\subsubsection{Real-Time Analytics and LLM Performance:}
The system monitors LLM modules using the \texttt{LearningGraph} class. It logs decisions and updates visual analytics for user feedback and system tuning~\cite{jiang2024importance}.

\paragraph{LLM Score Calculation.}
Scores are computed as follows:
\begin{equation}
J = \frac{|C \cap K|}{|C \cup K|}, \quad 
B = 0.6 + 0.4 \times J
\end{equation}

\paragraph{Bonus Adjustments.}
\begin{equation}
\mathrm{bonus}_{TP} = 0.1 \quad\text{if ``go'' or ``execute'' present}
\end{equation}
\begin{equation}
\mathrm{bonus}_{IR} = 0.15 \quad\text{if ``patrol'' present}
\end{equation}
\begin{equation}
\mathrm{bonus}_{MS} =
\begin{cases}
0.1, & \text{if modality matches},\\
-0.05, & \text{otherwise};
\end{cases}
\end{equation}
\begin{equation}
\mathrm{bonus}_{CG} = 0.1 \times f_{\mathrm{align}}(\text{context},\text{command})
\end{equation}

\paragraph{Final Score.}
\begin{equation}
S_{i}^{\text{new}} = B + \mathrm{bonus}_{i}, \quad 
S_{i} = \frac{S_{i}^{\text{new}} + w_{i}}{2}
\end{equation}

\paragraph{Weight Update~\cite{bottou2012stochastic}.}
\begin{equation}
w_{i} \leftarrow w_{i} + \eta\,(S_{i} - w_{i})
\end{equation}

\paragraph{Satisfaction Level.}
\begin{equation}
\text{Satisfaction} =
\begin{cases}
\text{Very High}, & c=3,\\
\text{High}, & c=2,\\
\text{Medium}, & c=1,\\
\text{Low}, & c=0.
\end{cases}
\end{equation}

\begin{figure}[!t]
  \centering
  \begin{minipage}[b]{0.48\textwidth}
    \centering
    \includegraphics[width=\textwidth]{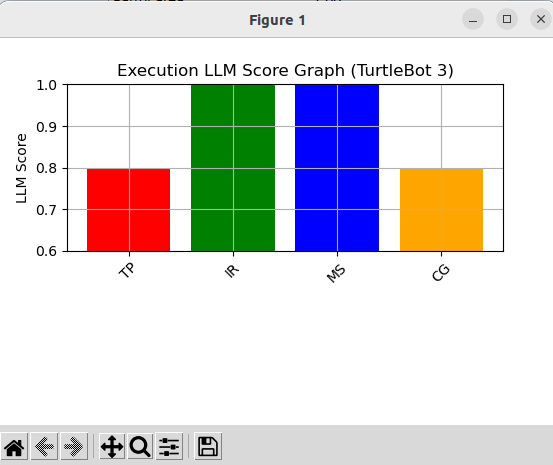}
    \caption*{(a) LLM score graph.}
    \label{fig:llm_score}
  \end{minipage}\hfill
  \begin{minipage}[b]{0.48\textwidth}
    \centering
    \includegraphics[width=\textwidth]{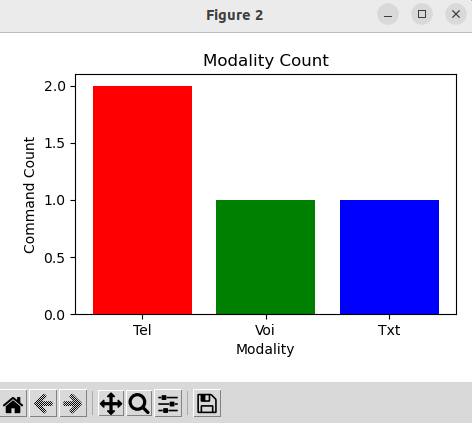}
    \caption*{(b) Modality count graph.}
    \label{fig:modality_count}
  \end{minipage}
  \caption{(a) LLM score graph; (b) Modality count graph.}
  \label{fig:performance_graphs}
\end{figure}

\paragraph{LLM Score Graph and Modality Count:}
These charts visualize module performance and usage frequency of each modality (Text, Voice, Teleop), as shown in Fig.~\ref{fig:performance_graphs}.

\subsection{Robot Swarm Layer:}
The Robot Swarm Layer receives and executes the refined, context-aware commands generated by the Processing Layer. This layer operates on the ROS2 framework and enables multiple robots (e.g., TurtleBot 1, TurtleBot 2, TurtleBot 3) to subscribe to the command topic, interpret the commands, and perform the required actions. Furthermore, each robot sends feedback on its execution status back to the central system, forming a closed feedback loop that enables continuous adaptation and performance monitoring.
\subsubsection{Command Reception and Parsing:}
Each robot in the swarm runs a ROS2 subscriber node that listens to a designated topic. When a command message is published, the following steps occur:
\begin{itemize}
    \item The robot\textquoteright s ROS2 node subscribes to the command topic. This is implemented using the ROS2 subscription mechanism.
    \item When a message is published, the subscriber node captures it.
    \item The received message, typically in JSON string format, is parsed to extract three key fields: the first one is the target identifier to identify which robot got the request to execute the command, the second one is the command to refine the published context, and the last one is the modality which specifies which modality should be processed. 
    \item The robot checks if the \enquote{target} field in the message matches its identifier. If it does, the robot processes the command; otherwise, the message is ignored.
    \item The users can also easily see which commands are sent (published) from the \emph{SwarmChat} and received (subscribed) by the TurtleBots from the buttons \enquote{Show Published Commands} and \enquote{Show Received Commands} respectively. 
\end{itemize}

\subsubsection{Command Interpretation and Execution:}
Once the correct robot receives the command, it interprets and executes by following these:

\paragraph{Command Interpretation:} The robot parses the command string. For instance:
\begin{itemize}
    \item If the modality is Text or Voice, the command is interpreted as a high-level instruction (e.g., ``Move forward and patrol the area'').
    \item In Teleop mode, the command may require further input (e.g., key presses such as ``P'', ``F'', ``B'', ``L'', ``R'', ``W'', ``A'', ``S'', and ``D'') etc.
    \item For customised commands, the robot also interprets and executes the customised context and modality given by the user which gives them much more flexibility outside the fixed contexts generated by the \emph{SwarmChat}. 
\end{itemize}

\paragraph{Conversion to Motion Commands:}
The robot converts the high-level instruction into low-level control signals, depending on the command. For example, if the command involves movement, it generates velocity commands (using ROS2 Twist messages) that control linear and angular velocities.

\begin{figure}
\centering
\includegraphics[width=\textwidth]{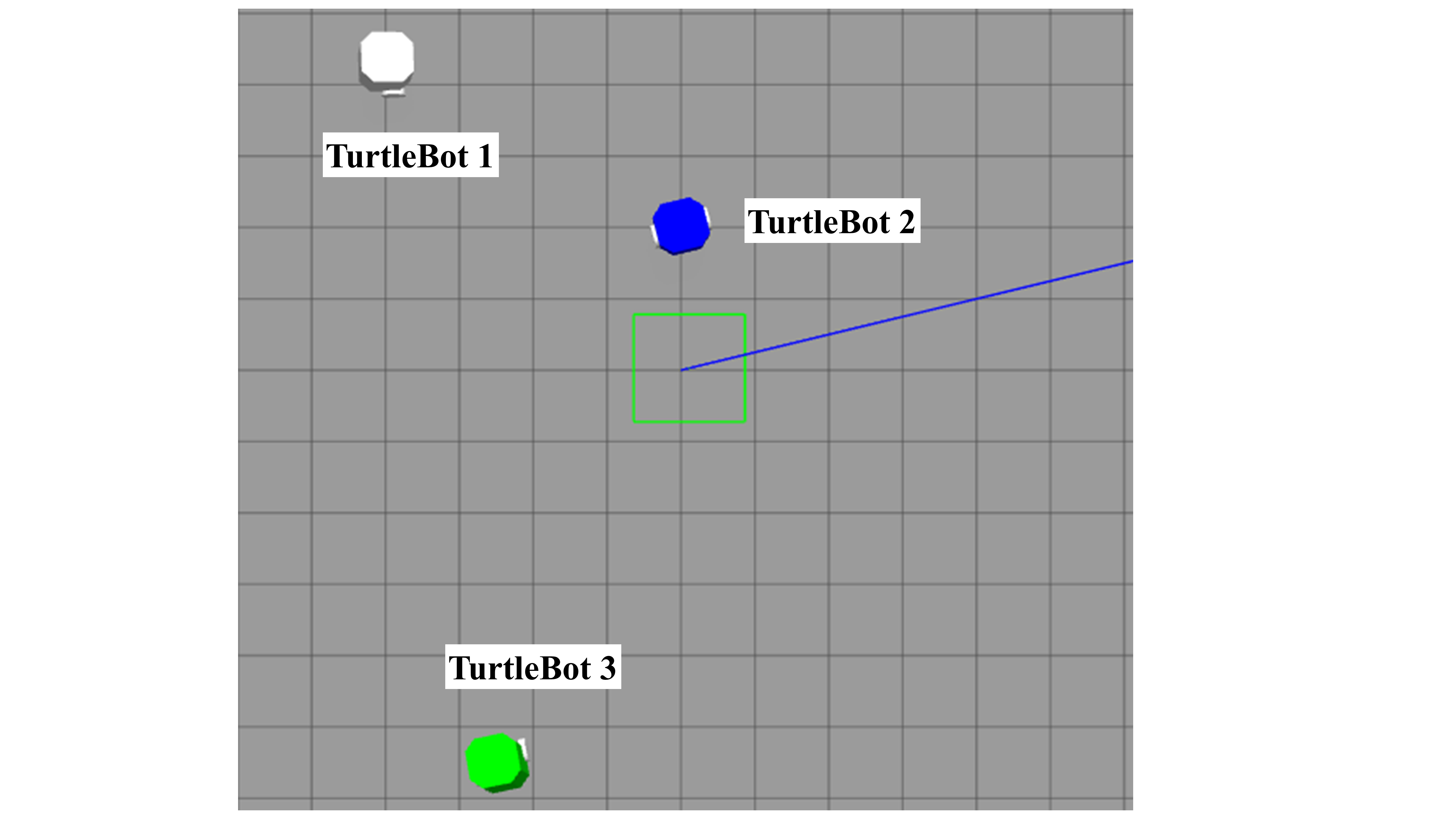}
\caption{Swarm movement with commands.}\label{robot}
\end{figure}

The robot swarm movement is shown in Fig.~\ref{robot}.

\subsubsection{Feedback Generation and Publication:}
After executing the command, each robot generates feedback to inform the central system about its current state. Using a dedicated ROS2 publisher, the robot sends this feedback message on a designated topic (e.g., swarmchat/feedback). The central system then uses the message to update real-time analytics and adjust future commands.

\section{Context Awareness and Cognitive Load Management}
The \emph{SwarmChat} can manage context awareness by dynamically generating relevant context based on given keywords and computing the context similarity score. This task improves the adaptivity of the interface. Moreover, in terms of cognitive load management, this interface aims to provide a less stressful user experience with multiple options. Users can easily add customised commands to their preferred modality for sending commands. If they choose \enquote{Voice} as their preferred modality and are not confident with the microphone, they can simply press \enquote{V} to send that voice command. They can also express their opinion based on the result of the customized commands. Additionally, users can select any context or modality outside of the suggested ones. They can track swarm actions and immediately receive acknowledgment of whether their command has been received or executed.

\section{System Evaluation and LLM Module Analysis}
\begin{figure}
\centering
\includegraphics[width=\textwidth]{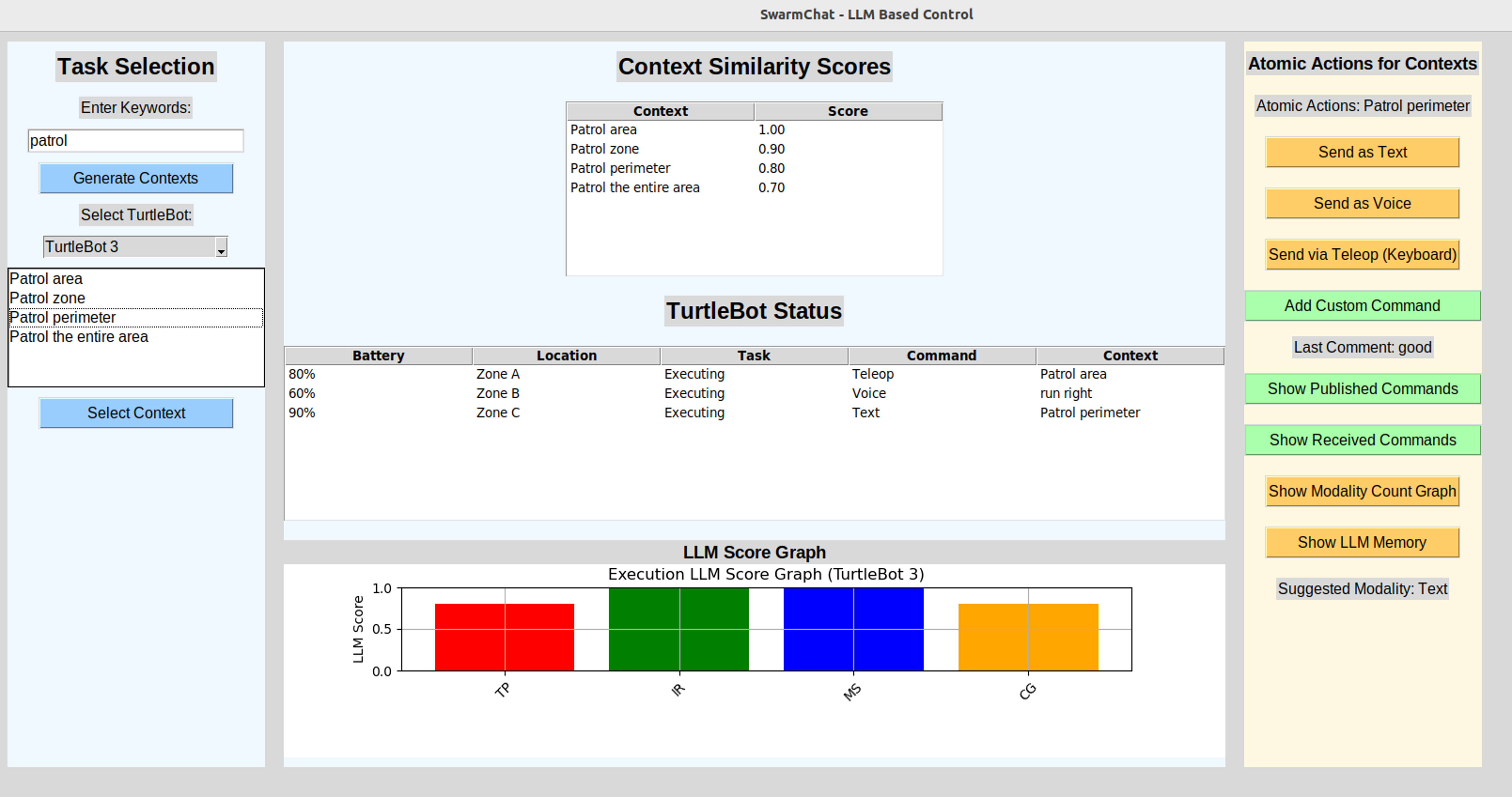}
\caption{Interface of developed SwarmChat.}\label{SC}
\end{figure}

To assess SwarmChat\textquoteright s performance, real-time interaction scenarios were conducted using various user inputs, modalities, and target robots. Table~\ref{tab:llm} presents a detailed scenario of the LLM module\textquoteright s performance during command execution. It highlights how each module, like Task Planner (TP), Intent Recognition (IR), Modality Selection (MS), and Context Generator (CG), processes commands, generates contexts, and selects modalities inside SwamrChat. The table includes performance scores, suggested vs. user-selected modalities, satisfaction levels, and user feedback for customised commands. The system interface supporting these actions is shown in Fig.~\ref{SC}.

\begin{table}
\caption{LLM Performance Table: Suggested (Sug.), Satisfaction (Sat.), and Comment (Com.)}\label{tab:llm}
\centering
\begin{tabular}{|l|l|l|l|l|l|l|l|}
\hline
LLM &  Context & Score & Sug. & User & Sat. & Decision & Com. \\
\hline
TP & Patrol area      & 1.00 & Teleop & Teleop (P)      & Very High & Execute ``Patrol area''      &      \\
TP & Patrol zone      & 1.00 & Teleop & Teleop (F)      & Very High & Execute ``Patrol zone''      &      \\
TP & run right        & 0.60 & Voice  & Voice           & Medium    & Execute ``run right''        & good \\
TP & Patrol perimeter & 1.00 & Text   & Text            & High      & Execute ``Patrol perimeter'' &      \\
\hline
IR & Patrol area      & 1.00 & Teleop & Teleop (P)      & Very High & Patrol mode activated      &      \\
IR & Patrol zone      & 1.00 & Teleop & Teleop (F)      & Very High & Patrol mode activated      &      \\
IR & run right        & 0.60 & Voice  & Voice           & Medium    & Patrol mode activated      & good \\
IR & Patrol perimeter & 1.00 & Text   & Text            & High      & Patrol mode activated      &      \\
\hline
MS & Patrol area      & 1.00 & Teleop & Teleop (P)      & Very High & Teleop                     &      \\
MS & Patrol zone      & 1.00 & Teleop & Teleop (F)      & Very High & Teleop                     &      \\
MS & run right        & 0.80 & Voice  & Voice           & Medium    & Text                       & good \\
MS & Patrol perimeter & 1.00 & Text   & Text            & High      & Text                       &      \\
\hline
CG & Patrol area      & 1.00 & Teleop & Teleop (P)      & Very High & Patrol area                &      \\
CG & Patrol zone      & 0.90 & Teleop & Teleop (F)      & Very High & Patrol area                &      \\
CG & run right        & 0.60 & Voice  & Voice           & Medium    & Patrol area                & good \\
CG & Patrol perimeter & 0.80 & Text   & Text            & High      & Patrol area                &      \\
\hline
\end{tabular}
\end{table}

\section{Conclusion \& Future Work}
In summary, the proposed \emph{SwarmChat} has made notable strides in enhancing human-swarm interaction by deploying an innovative multimodal chatbot system. This system effectively bridges the communication gap between humans and robotic swarms by leveraging advanced natural language processing and real-time analytics to create a multimodal interface. The interface is designed with the potential to alleviate user stress and facilitate the management of complex swarm operations. 
The current interface has some limitations. It relies on rule-based intent recognition, which may reduce adaptability to complex or ambiguous user inputs. Future work aims to address this limitation by developing more flexible, learning-based intent understanding methods. Additionally, the system's experimental evaluation is still in its initial stages; we have tested the system using various keyword combinations to assess the performance of the LLM models.
Future work will focus on expanding our experimental analysis by conducting large-scale user studies and direct comparisons with contemporary human-swarm interaction methods. The planned experiments aim to rigorously evaluate SwarmChat\textquoteright s effectiveness in diverse operational scenarios, supported by robust statistical validation of the reported outcomes.
We also plan to integrate in-depth statistical evaluations (such as ANOVA and t-tests) within the SwarmChat analytics dashboard to facilitate real-time comparison of user performance, feedback ratings, and system recommendations, thereby strengthening the empirical foundation for future improvements.
Furthermore, user study protocols will be expanded to capture subjective ratings (e.g., trust, clarity, comfort) after each interaction, objective task performance, and the degree of agreement between system recommendations and user choices, enabling more standardized evaluation of multimodal interaction.
Future developments will include integrating trainable large language models and performance evaluation with a diverse range of human users to more rigorously assess overall effectiveness and robustness in practical settings. The voice input modality will be further improved to better capture diverse accents and explore additional input methods. Finally, efforts will focus on optimizing computational requirements to reduce the system load on low-power devices, enabling broader deployment opportunities.

\bibliographystyle{splncs04}
\bibliography{main}
\end{document}